# Leaving Goals on the Pitch:
# Evaluating Decision Making in Soccer

Maaike Van Roy, Pieter Robberechts, Wen-Chi Yang, Luc De Raedt, Jesse Davis
KU Leuven, Department of Computer Science; Leuven.AI
{firstname.lastname}@kuleuven.be

Soccer Track - AMVC3NU3A

## 1. Introduction

"You can't score if you don't shoot." – Johan Cruyff

Analysis of the popular expected goals (xG) metric in soccer has determined that long-distance shots have a much lower chance of resulting in a goal than closer ones. More crucially, it allowed us to realize that a (slightly) smaller number of high-quality attempts will likely yield more goals than a slew of low-quality ones. This has driven a change in teams' shooting policies in terms of shot volumes and locations, which is illustrated in Figure 1 for the English Premier League (EPL). One sees a slight decline in the number of shot attempts per match and a larger decline of around 20% in the number of shots from distance per season from the 2013/14 to the 2018/19 season.

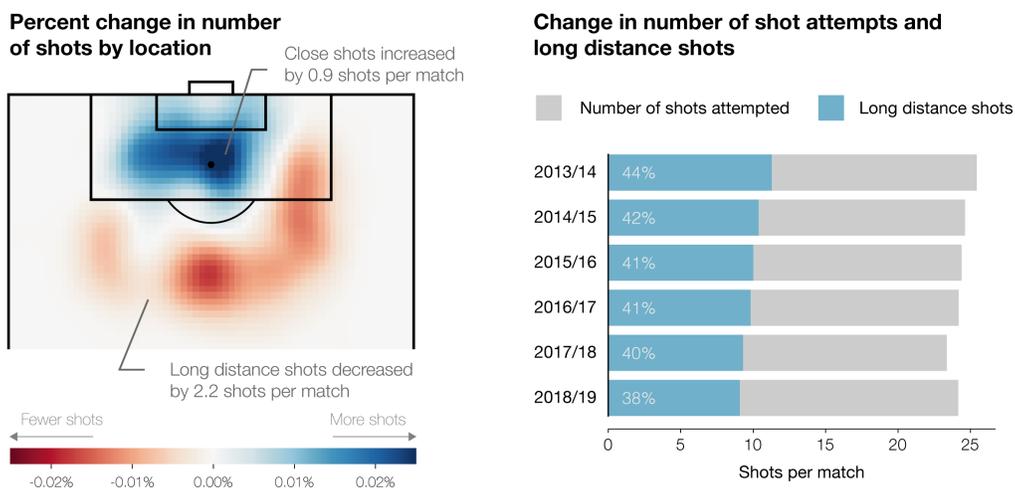

**Figure 1:** Evolution of shooting in the English Premier League between 2013/14 and 2018/19. A long-distance shot is defined as a shot taken from outside the penalty box. Over the course of this time period, the number of these shots has declined by around 20%.

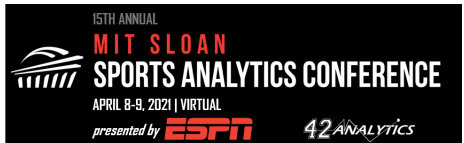

Despite the observable changes in shooting behavior, Cruyff's adage lives on in a small group of high-volume long-distance shooters such as Christian Eriksen, Paul Pogba, Harry Kane, Kevin De Bruyne, Heung-Min Son, Eden Hazard, and Gylfi Sigurdsson. All these players attempted more than 45 shots across the 2017/18 and 2018/19 EPL from the long-distance zone shown in Figure 2. In aggregate, they converted 6.5% of their 416 long-distance shots.[1] In the 4455 times these seven players touched the ball in the long-distance zone and did not shoot, their teams only ended up scoring a goal in 2.1% of these possessions.

This paper attempts to reconcile the insights from xG with the aforementioned statistics by using a nuanced approach to analyze decision making and policies around shooting in soccer. There is a risk-reward trade-off to forgoing a shot, which goes back to Cruyff's quote: you can't score if you don't shoot. If a player shoots immediately, this ensures at least the chance of scoring. The potential payoff of not shooting is that an even better shot may arise down the line, but there is no guarantee of this happening as the team may lose the ball prior to another shooting chance arising. This leaves players and coaches to ponder concrete questions such as:

1. Given that a player possesses the ball in a specific location, what is the chance of generating a better shot later on in the possession?
2. A player possesses the ball several meters outside the penalty box, he or she has the option to pass the ball to a teammate on the flank or shoot: What should he or she do?
3. How would a 10% increase or decrease in a team's probability of shooting from long-distance affect the number of goals the team would be expected to score over the course of the season?

Unfortunately, these are difficult questions to answer. On the one hand, one must reason about the number of ways a possession could play out as well as the inherent uncertainty of actions succeeding (e.g., a player mishitting the ball). On the other hand, understanding the effect of different shooting policies requires *counterfactual* reasoning: one needs to reason about *what* could have happened *if* a different policy had been followed [8].

This paper proposes a novel generic framework to reason about questions such as these in soccer by combining techniques from machine learning and artificial intelligence (AI). We model how a team has behaved offensively over the course of two seasons by learning a Markov Decision Process (MDP) from event stream data. This type of data annotates all touches of the ball such as passes, dribbles and shots that occur during a match. Specifically, we model the probability that a team will attempt to move the ball to a specific location or shoot based on the current location of the ball. Then, we apply techniques from AI to reason about the learned MDP. First, we use techniques from probabilistic verification to analyze a team's chance of scoring if they would shoot from a specific state versus their *chance of scoring in the future* if they would forgo a shot. Second, we explore the effects of modifying a team's shooting policy (i.e., how often they shoot versus move) from various locations on the total number of goals that the team would be expected to score over the course of a season.

---

[1] This number is likely an underestimate as we did not take rebounds into account (i.e., other chances may have been created due to a shot from distance).

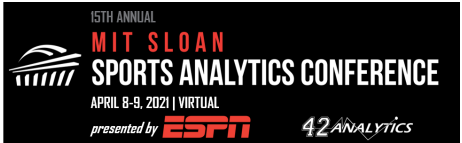

We use our framework to analyze the aforementioned questions about shooting in specific situations as well as the effect of altering a team's shooting policy over the course of a season. Concretely, we find:

- Teams are likely taking too few long-distance shots and are likely leaving goals on the pitch.
- Per team, we identify specific long-distance areas on the pitch where increasing their shot frequency by 10-20% would likely lead to an increase of 0.5-1.5 goals scored in a season.
- There are a number of team-specific locations within the long-distance zone from which even very strong teams like Chelsea are unlikely to generate a better shot later in the possession. Hence, if a team has a viable shot from these locations, they should take it.
- Relying on multiple movement actions (i.e., passes or carries) to generate a shot, particularly if the first move entails passing the ball to the flank, increases the risk of not generating a future shot.

These insights are relevant for a number of use cases within soccer clubs. First, managers can use them to inform tactics such as helping to specify how often and from what locations they should encourage players to shoot. Second, they can be applied during training to help guide a player's decision making in terms of in what situations moving or shooting is appropriate. Third, it may help with match preparation in terms of identifying their opponent's strengths and tendencies, which could be useful for tasks such as line-up selection.

## 2. Modeling Team Behavior

Our goal is to model the offensive behavior of players on a specific team during a soccer match in terms of which actions they tend to perform in various locations of the pitch. A natural formalism to model this behavior is to use Markov Decision Processes (MDPs). This well-known formalism models how an agent behaves in a dynamic environment and how the environment transitions between different states based on which action the agent decides to perform. Our MDP works on the level of a possession, which is a contiguous sequence of events where the same team possesses the ball. At a high level, our MDP models the probability that a player possessing the ball will attempt to either (1) move the ball, e.g., by passing it to a teammate in another location on the pitch, or (2) shoot. The probability of selecting a particular action solely depends on the location of the player currently possessing the ball. Next, we formally define our MDP and describe how to learn it from data.

### 2.1. Defining the Markov decision process
An MDP consists of the following five components:

- **the state space**, which describes the various states in the environment;
- **the action space**, which specifies which actions are possible;
- **the transition function**, which specifies the probability of transitioning between any two states;
- **the policy**, which determines the probability of selecting actions in states; and
- **the reward function**, which determines the immediate reward received for each transition.

Next, we define each of these components.

**State space:** We split our set of possible states into a set of *field* states and a set of absorbing states. The field states are those states that teams can enter and exit during a possession sequence. These states are defined by partitioning the field into a number of disjoint zones. The current game state is defined by the current location of the ball. Having a more fine-grained partitioning is more informative as it allows for distinguishing between subtly different situations. Unfortunately, it also poses issues in terms of data sparsity as there must be a sufficiently large number of actions in each zone to enable learning the transition function and policy. Therefore, we select an appropriate state space depending on our use case of interest: shooting. Because shooting (mainly) only occurs on the offensive half of the field, we partition this part of the field into states by overlaying a 22-by-17 grid and use a single state for the defensive half of the field. Figure 2 provides an illustration of the fine-grained set of field states used.

We define three absorbing states that denote the end of a possession sequence: $s_{goal}$, $s_{no\_goal}$, and $s_{loss\_possession}$. These denote a goal, a missed shot, and the loss of possession after a move action, respectively. Once one of these states has been reached, the possession sequence ends. Note that recovering a rebound from a missed shot or the ball going out of bounds would constitute a new possession.

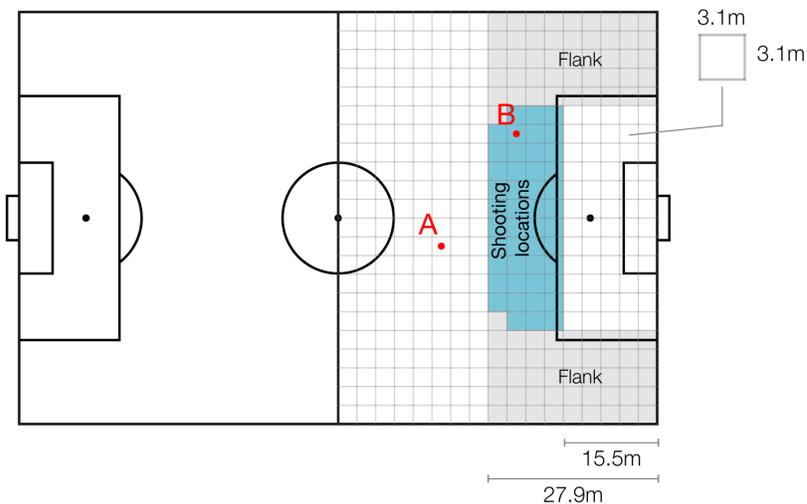

**Figure 2:** Illustration of the 375 field states used. The long-distance shooting locations are shown in blue. Our goal is to analyze behavior related to shooting; hence, we use a fine-grained 22-by-17 partitioning in the offensive half and a single state for the defensive half.

**Action space:** The set of possible actions includes attempting to move the ball to a specific location (i.e., state) on the field, and shooting. An example of a move action is attempting to move the ball from zone A to zone B in Figure 2. The move action does not differentiate among the various ways (e.g., pass, cross, carry) that a player can use to move the ball. Each action can either succeed or fail. An example of a successful move action is a pass where a teammate was able to receive and control the ball. Examples of a failed move action include an intercepted pass or a pass that goes out of bounds. A successful shot is one that results in a goal. An example of a failed shot is one that is saved by the goalie.

**Transition function:** This function specifies the probability of transitioning between every possible pair of states. For all three absorbing states, the only possible transition is a self-loop with a probability of one. By definition, any other outgoing transition is impossible and receives a probability of zero. For all field states *s*, the transition function is defined as follows:

$$P(s, a, s') = Pr[S_{t+1} = s' \mid S_t = s, A_t = a]$$

where *t* denotes the time-step. This specifies that the next state only depends on the current state and the action that was chosen in the current state. That is, it does not consider any information about the past, which is also called the Markov property. Concretely, the transition function is defined as:

- $P(s, move\_to(s'), s')$ is the probability of successfully moving the ball from *s* to *s'*, where *s'* is always another field state.
- $P(s, move\_to(s'), s_{loss\_possession}) = 1 - P(s, move\_to(s'), s')$. This captures the probability of the move action failing and the system transitioning to the absorbing "loss-of-possession" state.
- $P(s, shoot, s_{goal})$ is the probability of successfully scoring a goal from field state *s*. This can be viewed as a location-based xG value.
- $P(s, shoot, s_{no\_goal}) = 1 - P(s, shoot, s_{goal})$. This is the probability of missing the shot, and the system transitioning to the absorbing "no-goal" state.
- Any other combination results in a transition probability of zero.

**Policy:** The policy defines the probability that a team will decide to perform a particular action in a given state. In our case, it specifies the probability of choosing an action *a* in a field state *s*:

$$\pi(a \mid s) = Pr[A_t = a \mid S_t = s]$$

The policy is the only part of the MDP that the team (player) can immediately control. This is then also the perfect starting point for answering counterfactual questions. The policy can be adapted to a certain "What if"-scenario (e.g., performing certain actions more than usual), and the effects of these changes can be observed. The resulting insights could aid teams in their decision making.

**Reward function:** The reward function assigns rewards to various transitions in the MDP. Our reward function assigns a reward of 1 when a goal is scored from one of the field states. All other transitions receive a reward of 0.

## 2.2. Learning the MDP from data

Both the transition function and the policy of the MDP need to be learned from event data. For this, we use the 2017/18 and 2018/19 English Premier League event stream data, which has been provided by StatsBomb.[2] We learn one MDP per team and only consider the 17 teams that appeared in both these seasons. As each team plays every other team twice during each season, with a total of 20 teams in the league, this results in data from a total of 76 matches per team.

Generally, learning the MDP is done by using the observed transitions in the data to estimate the probabilities:

---

[2] https://statsbomb.com/

$$P(s, move\_to(s'), s') = c^+_{s,s'}/c_{s,s'}$$

$$\pi(move\_to(s') \mid s) = c_{s,s'}/c_s$$

$$P(s, shoot, s_{goal}) = c^+_{s,s_{goal}}/c_{s,s_{goal}}$$

$$\pi(shoot \mid s) = c_{s,s_{goal}}/c_s$$

Here $c_{s,s'}$ is the number of observed actions that start in *s* with the intent of moving the ball to state *s'*, $c^+_{s,s'}$ is the number of those actions that successfully reach state *s'*. The quantity $c_{s,s_{goal}}$ is the number of observed shots in state *s*, and $c^+_{s,s_{goal}}$ is the number of shots in state *s* that result in a goal. Finally, $c_s$ is the total number of observed actions that start in *s*.

The choice for an action space that includes the intention of actions complicates estimating these probabilities. In particular, $c_{s,s'}$ cannot be directly obtained from the data. To estimate this number, both the successful and unsuccessful move actions from state *s* to *s'* need to be known. However, for unsuccessful actions, the intended end location is unknown. For example, consider a pass that goes out of bounds. Event stream data records the location where the ball went out of bounds, but not the location it was intended to reach. Hence, it is not possible to estimate the number of unsuccessful move actions from state *s* that were intended to reach state *s'* directly from the data.

To solve this problem, we use Gradient Boosted Trees Ensembles to predict the intended end location of actions based on the characteristics of the actions and what has happened prior to the actions. This allows us to estimate the probabilities needed to finalize the construction of the MDP. The evaluation of how accurately our MDP models the data can be found in Appendix A.1.

## 3. Evaluating Decision Making with Counterfactual Reasoning

An MDP learned from data captures a team's observed policy in terms of the team's probability of selecting a certain action in each field state. The key question is then how to reason about the policy in order to provide insights that coaches and players can exploit. Consider the following broad classes of questions:

1. **What** is the expected outcome of taking a particular action or sequence of actions?
2. **What** is the relative merit of two particular policies in a specific location?
3. **What** would happen **if** a team changed from one policy to another?

Our framework employs two techniques to reason about decision making using the MDP. The first technique aims to answer the "**What**" questions. It compares the possible outcomes of two (or more) possible actions in a specific game situation. For example, would shooting or first playing the ball wide and then crossing into the box result in a higher chance of scoring? The second technique aims to answer the "**What if**" question. It describes how to modify a team's policy and assess the effects of the change. For example, this could be used to assess the impact of generally shooting more or less often from outside the penalty box.

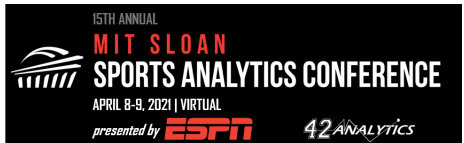

## 3.1. What should players decide to do?

During the course of a game, players must continually make decisions about which actions to perform. Evaluating the potential benefits of different courses of actions is non-trivial. While a player may have a good sense about the chance of scoring from a particular shot or the chance of successfully completing a pass, it is harder to assess the probability of generating a shot within the next three actions. Assessing this chance involves accounting for the uncertainty about what actions may occur in the future and whether these actions will succeed or fail. From a computational perspective, this is also a complicated task. This would be (nearly) impossible to do based solely on the raw event stream data. However, when using an MDP to model a team's behavior, this becomes possible by exploiting techniques from the field of AI.

More specifically, the AI field of probabilistic model checking has developed powerful tools such as PRISM [7] and STORM [5] for reasoning about the probability of certain behaviors arising in a stochastic system like an MDP. Among other capabilities, these tools can exactly compute the probability of certain properties arising in the model. A simple example of a property could be checking the probability that a sequence starting in state A (Figure 2) will reach the penalty box. It is also possible to reason about properties that place more complex constraints on the actions taken. For example, a complex property would entail checking the probability of scoring from a sequence that starts in the blue area in Figure 2, after executing a move action to one of the flanks (grey in Figure 2), followed by another move action and then ends with a shot. In a nutshell, these tools compute these probabilities by checking the properties against all possible behaviors (i.e., sequences of actions) that can arise in the system. Characterizing the expressiveness of these model checkers and how to formulate these properties is beyond the scope of this paper, see the PRISM [20] or STORM [21] websites for a tutorial.

## 3.2. What if a team decided to alter its policy?

The goal of our second technique is to assess the effects of a team changing its policy. That is, what happens if changes are made to the probability of selecting various actions in a particular state. We evaluate the effect of changing the policy on the number of goals a team would be expected to score over the course of a season. However, this is non-trivial because changing the policy will affect both the number of shots a team will perform in each location and the success probability of the shots taken because there is likely a quantity-quality trade-off. For example, increasing the shot volume from a particular location would likely entail a team taking more low-quality shots from that location. Both the quantity and quality of the shots need to be considered when evaluating the effect of changing the policy.

### 3.2.1 Changing the policy

We employ a relatively straightforward approach to altering a team's policy. We assume that when the probability of performing an action increases (decreases) in a state, the probability of all other actions decreases (increases) proportional to their original share. For example, consider the simple policy shown in Figure 3a. When increasing the probability of shooting by 50% (i.e., an additional probability of 0.1), the total probability of moving needs to decrease by 0.1 as a consequence. Decreasing the probability of the move actions according to their original share means that the probability of choosing to move to A becomes 0.3/0.8 * 0.7 and the probability of choosing to move to B becomes 0.5/0.8 * 0.7. This yields the modified policy shown in Figure 3b. The general formulas for modifying the policy are given in Appendix A.2.

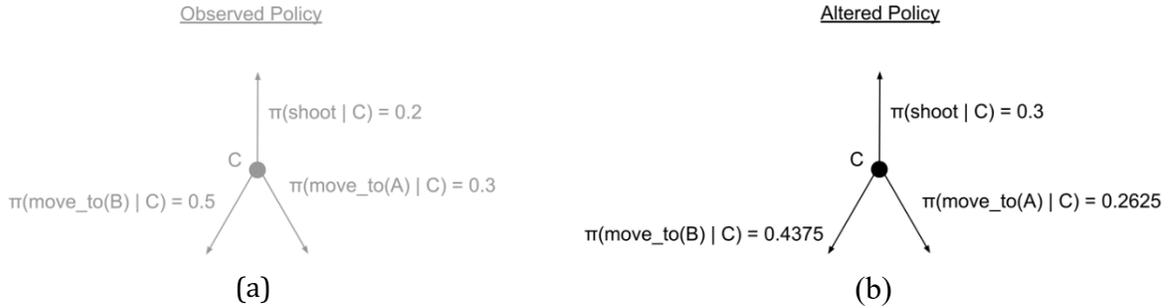

**Figure 3:** Illustration of changing the policy of a state C with three actions (i.e., shoot, move to state A, or move to state B). (a) The observed policy. (b) The altered policy where the probability of shooting is increased by 50%.

Changing the policy can be done either locally, by only changing the policy in one state, or more globally, by changing the policy in a (related) set of states. This same basic idea is also applicable for increasing or decreasing the probability of performing a move action. However, our main focus in this paper is to analyze policies around shooting and hence we do not explore directly manipulating these probabilities. This way of modifying a policy has been explored previously for basketball [14], [15].

### 3.2.2 Evaluating the effectiveness of a policy

To assess the effectiveness of a policy, we estimate the number of goals a team is expected to score over a season for this given policy. This can be done by knowing (1) the expected number of shots from each field state over the course of a season, and (2) the probability of a shot resulting in a goal for each field state. Formally, this can be stated as:

$$E[goals \mid \pi] = \sum_{s \in field\ states} E[shots\ in\ s \mid \pi] * P(s, shoot, s_{s\_goal})$$

Estimating the expected number of shots is relatively straightforward. Estimating $P(s, shoot, s_{s\_goal})$ for the observed policy $\pi$ is also trivial as it is simply the probability learned from the data. However, when we modify the policy to force a team to shoot more or less, this could affect the quality of the shots taken and hence their probability of resulting in a goal. If a team decides to shoot less, it stands to reason that they continue to take high-quality chances while trying to excise the lower-quality shots from their profile. The opposite holds when increasing the shot volume. The relationship between the frequency of a given shot type and its probability of succeeding has been discussed in basketball [3], [14], [15], and it seems reasonable to assume that such a relationship would also exist for soccer.

**Estimating $E[shots\ in\ s \mid \pi]$**

We estimate the expected number of shots a team will take in a particular field state $s$ as:

$$E[shots\ in\ s \mid \pi] = \pi(shoot \mid s) * E[visits\ to\ s \mid \pi]$$

Estimating the expected number of visits to a state over the course of a season can be computed by knowing (1) how many possessions start in each field state $s'$ and (2) given that a possession starts in $s'$, how many times field state $s$ will be visited prior to absorption. We compute the number of possessions starting in each field state *s'* by counting how often a possession starts from *s'* in the event stream data. We presume that this number is not affected by the policy. Fortunately, (2) can be directly derived from the MDP by computing its fundamental matrix *N*. The entry $N[i,j]$ contains the expected number of times field state *j* will be visited before an absorbing state is reached given that the possession sequence started in field state *i*. More formally, *N* can be computed as:

$$N = (I - Q)^{-1}$$

where I is the *f*-by-*f* identity matrix with *f* the number of field states and *Q* describes the probability of transitioning from one field state to another when fixing the policy. More precisely, each entry $Q[i,j]$ can be computed as $Q[i,j] = P(i, move\_to(j), j) * \pi(move\_to(j) | i)$. This policy $\pi$ can either be the observed or adjusted policy.

### Estimating $P'(s, shoot, s_{goal})$ in $\pi'$

Accurately modeling the frequency versus efficiency trade-off would be a complicated undertaking in its own right. Therefore, we approximate this trade-off using a simpler approach. Recall that $P(s, shoot, s_{s\_goal})$ is the xG for any shot taken in *s* because our model uses a purely location-based xG score. To obtain a fine-grained distribution over possible xG values for a given state *s* we use the xG values that StatsBomb awards to each shot in *s*. We then use the distribution of these xG values to derive an adjustment to the estimate of $P'(s, shoot, s_{s\_goal})$ for a modified policy $\pi'$. We apply a different modification for increasing and decreasing the number of shots.

Case 1: Increasing the propensity to shoot. In this case, each additional shot taken will likely be of a lower quality. In the new policy, $E[shots\ in\ s\ |\ \pi'] - E[shots\ in\ s\ |\ \pi]$ more shots will be taken than in the original policy. We only modify the xG value for the additional shots taken in $\pi'$. These shots are awarded an xG of

$$P(s, shoot, s_{s\_goal}) - (\mu_s - \mu_s^{low})$$

where (a) $\mu_s$ is the average StatsBomb xG of all shots occurring in field state *s* and (b) $\mu_s^{low}$ is the average StatsBomb xG of the below-average shots occurring in field state *s*. An illustration of the StatsBomb xG distribution for one field state including both $\mu_s$ and $\mu_s^{low}$ is shown in Figure 4.

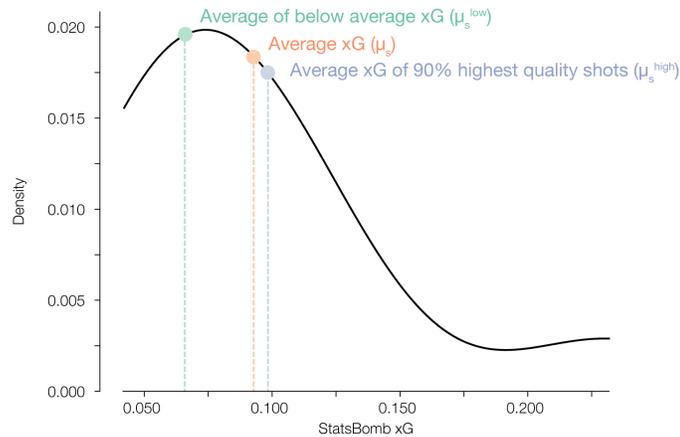

**StatsBomb xG distribution for one field state**

**Figure 4:** Distribution of the possible StatsBomb xG values in one field state. The orange line indicates the average xG value ($\mu_s$), the green line indicates the average of the below-average xG values ($\mu_s^{low}$), and the purple line indicates the average xG value of the shots with an xG value in the top 90% of all shots in that zone ($\mu_s^{high}$).

Case 2: Decreasing the propensity to shoot. In this case, fewer shots will be taken than in the original policy. Hence, the xG estimate for shooting in each state is too low. We counter this by modifying the xG values in each state where shooting has decreased. These shots are awarded an xG of

$$P(s, shoot, s_{goal}) + (\mu_s^{high} - \mu_s)$$

where (a) $\mu_s$ is again the average StatsBomb xG of all shots occurring in field state *s* and (b) $\mu_s^{high}$ is the average StatsBomb xG of the 1-*x* highest-quality shots occurring in field state *s*, when decreasing the propensity to shoot by *x*%. Thus, we drop the *x*% lowest-quality shots and compute the average xG of the remaining shots. Figure 4 illustrates $\mu_s^{high}$ for *x* equal to 10%.

**Accuracy of estimating the number of goals**
To assess the accuracy of our estimates, we calculate the relative error between (1) the estimated number of goals scored according to our analysis of the observed policy and (2) the actual number of goals recorded in the data for each team. The resulting average relative error over all teams is 11.38%. Given that an average team scores around 50 goals per season, this corresponds to roughly six goals. This indicates that our method produces fairly good estimates of the number of goals for the observed policy. We can assume that this will also be the case for the altered policy.

## 4. Use Cases

Our framework enables answering a multitude of (counterfactual) questions that are on the minds of players, coaches and fans. In this section, we present four concrete use cases. The first use case aims to investigate whether players should forgo a shot outside the penalty box (e.g., by passing it to a teammate) or take it. The second use case complements the first one by looking further into the future and aims to evaluate the odds of the team ever generating a better shooting chance than the one they have now. The third use case aims to reason about the possible effects of increasing or decreasing the number of long-distance shots on the expected number of goals a team would score in a season. The fourth use case combines both methodologies introduced in this paper to reason about the possible effects of a *targeted* increase or decrease in the number of long-distance shots on a team's expected number of goals in a season.

### 4.1. Should you pass up on the shot, or take it?

Consider a player possessing the ball a few meters outside the penalty box with two choices:

1. Take the shot now from the present location;
2. Try to move the ball to another location (e.g., pass it to a teammate) and shoot from the new location.

What decision should the player make? Common sense dictates that the best option is the one that has the highest chance of generating a goal. However, knowing which choice is best according to this criterion involves a trade-off. Taking the shot now ensures the possibility of scoring the goal. Forgoing the shot may generate a better opportunity down the line, but does entail some risk. The pass may be errant or your teammate may miscontrol it, leading to the shooting opportunity evaporating. While the best decision will clearly be context-specific (e.g., how much pressure is the shooter under, where is the teammate located), we can use the techniques from Section 3.1 to get a general sense of when it is preferable to shoot and when it is better to pass.

First, we consider two specific movement conditions: (1) there is exactly one attempted move action after which a shot is attempted, and (2) there are exactly two move actions to any field location after which a shot is attempted. We use PRISM [7], a verification tool for Markov models, to compute the exact probability of scoring in both scenarios according to the learned MDPs for a sequence of play beginning in each of the states inside the blue colored box in Figure 2.

Figure 5 visualizes for each move scenario the difference in xG between that move scenario and directly shooting for four teams: Chelsea, Everton, Huddersfield Town, and Manchester United. Red indicates where immediately shooting is the better choice as moving would decrease your odds of scoring. Blue indicates when moving would optimize your chances of scoring. The top considers the first scenario of exactly one move action prior to shooting and the bottom shows the results for performing exactly two move actions to any field location prior to shooting.

These results show that for each team there are zones where moving is better and also zones where shooting is better. Interestingly, the differences in benefits of shooting and moving are not symmetric. When shooting is preferred, the payoff in terms of the increase in the chance of scoring is much higher than when moving is preferred. The locations where shooting is preferred vary by team, but commonalities do arise. Zones that are in the front of the center of the goal are generally regarded as good locations to shoot from. Intuitively, this makes sense because when a player attempts a forward central pass from this location, the receiving player almost always needs to turn before being able to shoot. Thus, immediately shooting is a better choice than a pass followed by a subsequent shot. Additionally, locations that lie to the left of this area can also be spotted as good shooting locations for some teams. One possible explanation could be the use of inverted wingers, like Eden Hazard during his seasons at Chelsea and Everton's Yannick Bolasie. These players can cut infield to get the ball on their dominant foot, which facilitates shooting.

When increasing the number of required move actions prior to shooting, the benefit of moving decreases. Hence, in some zones shooting is now preferable whereas moving was better in the one-move scenario. This is most clearly visible at the border of the penalty box. In these states, performing more move actions increases the odds of losing the ball and consequently missing out on the chance of scoring. Concretely, it is preferable to take an available shot in these zones, unless you can move directly (i.e., in one step) to a player who will be able to shoot. The figures for all teams can be found in Appendix A.3.

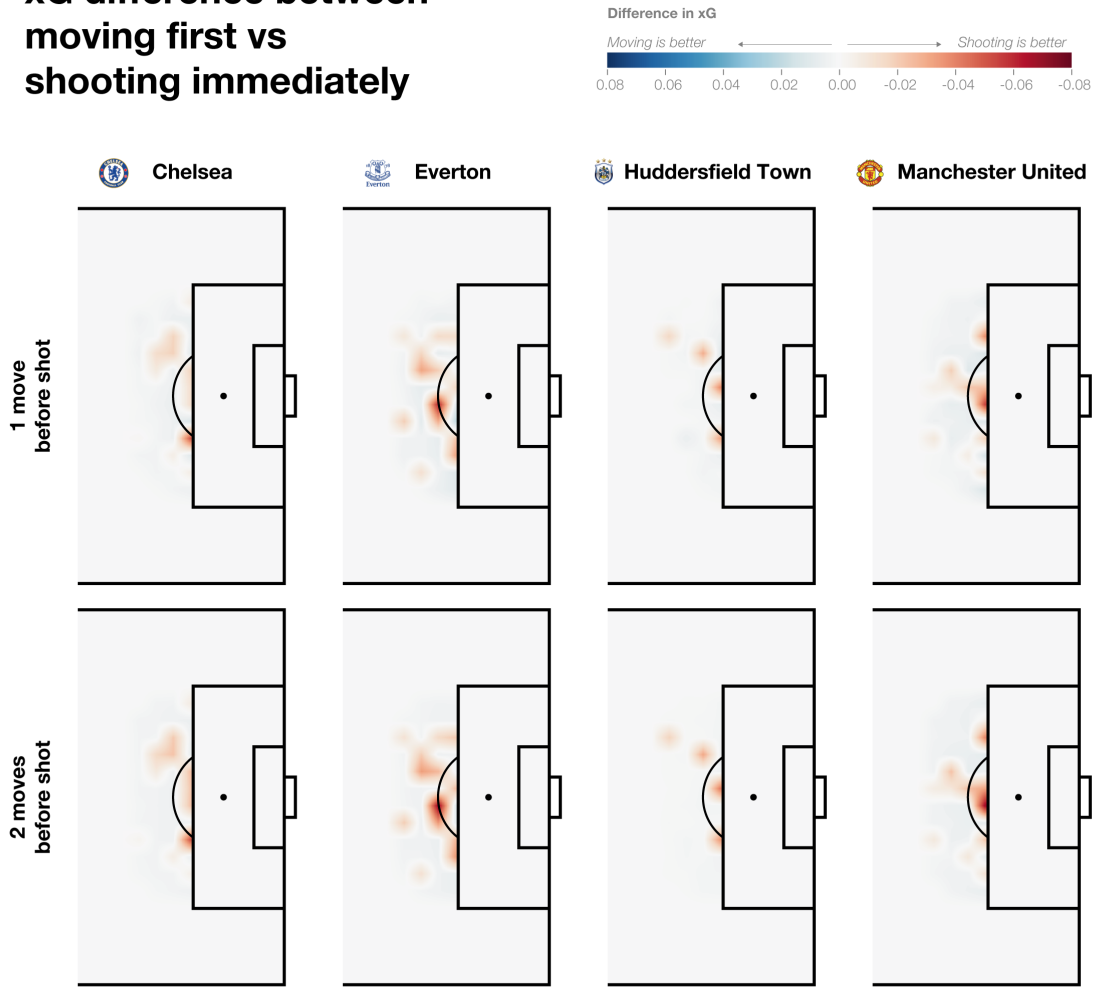

**Figure 5:** The difference in xG between moving once prior to shooting (top row) or moving exactly twice to any location prior to shooting (bottom row) versus directly shooting. Red indicates where immediately shooting is the better choice as moving would decrease your odds of scoring. Blue indicates when moving would optimize your chances of scoring. The best choice in each zone is shown for Chelsea, Everton, Huddersfield Town and Manchester United. The locations where shooting (moving) is the better choice vary from team to team. However, a common area where it is beneficial to shoot for all teams is the border of the penalty box. When adding yet another move action to the sequence, the benefit of moving starts to disappear in certain states. This is clearly visible on the border of the penalty box.

Now, we will further restrict the second move scenario by requiring that the first action in the sequence must move the ball to one of the flanks, which are defined as the gray-shaded regions in Figure 2. Often the flanks are used to open up space in the hopes of generating a better shot later in the possession sequence. This analysis also highlights the power and flexibility of probabilistic model checking to incorporate and reason about various constraints on an MDP. Figure 6 visualizes the difference in xG between this scenario and directly shooting for Chelsea and Everton. Imposing this additional restriction of first moving the ball to the flank decreases the chance of generating a goal later on and hence makes shooting more advantageous in many areas.

These types of analysis and the insights that they provide can aid teams in several ways. Offensively, it can help teams provide concrete advice to players about the desired decisions to make in a particular part of the pitch when specific situations arise. Defensively, it gives teams an indication of how an opponent should behave, which they can use to help craft a plan to disrupt them. For example, when deciding upon the line-up for a game, a team playing against Everton might want to use a holding defensive midfielder with the aim of clogging the area to the left of the penalty arc in order to force them to move the ball to the flank instead of allowing them to shoot.

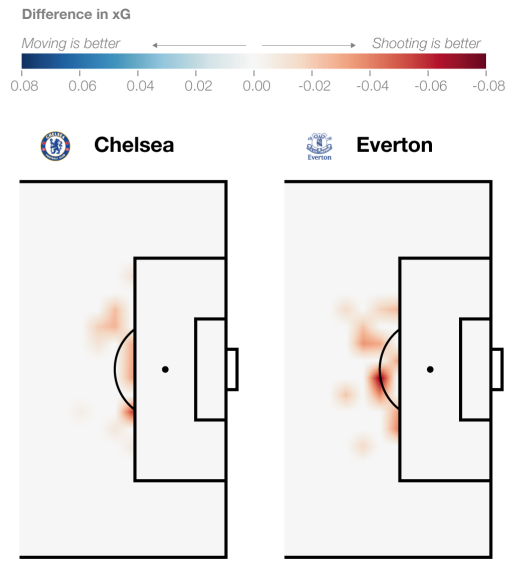

**Figure 6:** The difference in xG between moving twice prior to shooting with the constraint that the first move action must move the ball to one of the flanks versus directly shooting. Red indicates where directly shooting is the better choice, blue indicates where moving would optimize your chances of scoring. The best choice for each zone is shown for Chelsea and Everton. Compared to the other move scenarios, there are more zones where shooting is preferred to moving. Moreover, the advantage of shooting over moving has increased.

## 4.2. Probability of generating a better shot later in the possession

To complement the analysis in the prior section, we look further into the future and answer the question: "For a specific location, what is the probability of ever generating a better shot than the one they have when outside the box?"

To answer this question, for each field state *s*, we use PRISM to compute the probability of generating a shot with a higher xG value than the xG value associated with shooting from *s*. Figure 7 shows these state-specific probabilities for the same teams as in the previous use cases. Our method identifies various regions with higher and lower odds of generating a better shot, which mostly correspond to the same areas identified in the analysis in Section 4.1. Interestingly, there are many locations where teams are very unlikely to generate a better shot. In fact, the probability of ever generating a better shot can be as low as 5% for some locations. An example of such a location for Chelsea is the region to the left of the penalty arc outside the box. Similarly, when a team is on the edge of the box, directly in front of the goal, the chances of generating a better shot are only between 5% and 10%. In other locations, like Manchester United's left side of attack, the probabilities can be quite high, at around 20%. Interestingly, the magnitude of the probabilities can vary substantially from team to team. For example, Huddersfield are very unlikely to ever generate a better shot. The figures for all other teams can be found in Appendix A.4.

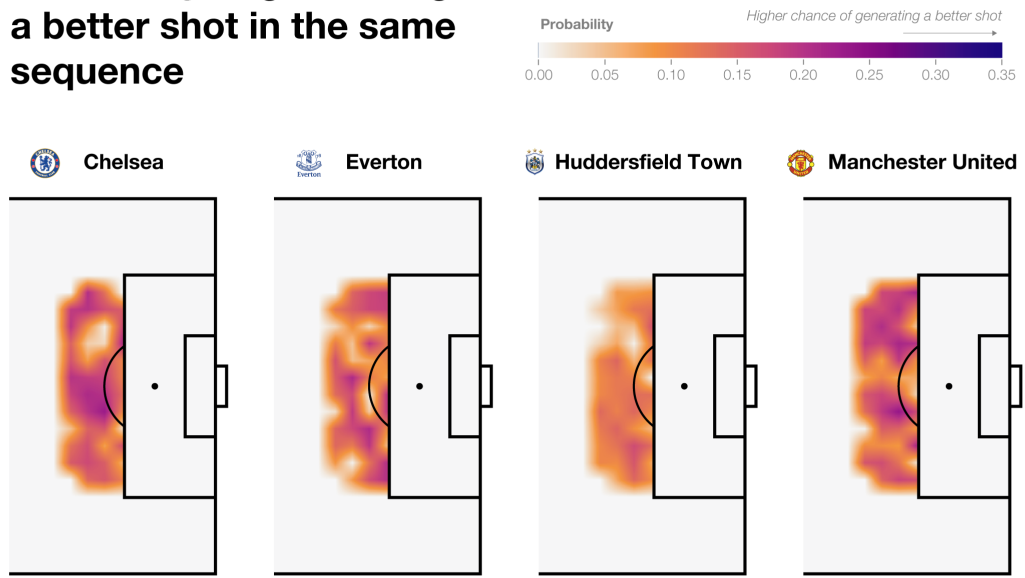

**Figure 7:** The probability of ever generating a better shot in the same possession sequence for Chelsea, Everton, Huddersfield Town, and Manchester United. The regions with the lowest probabilities correspond to the zones where shooting was preferable to moving in the previous use case. The magnitudes of these probabilities vary by team. Regardless of location, Huddersfield has a relatively small chance of ever creating a better-valued shot.

## 4.3. Uniformly shooting more or less from distance

Earlier this year, Toronto FC's Director of Analytics Devin Pleuler posed the question "What happens if players shot from distance 10% less frequently?" on Twitter.[3] Addressing this query complements the analysis in Section 4.1, which explores relatively narrow situations, by offering a more global assessment on the effect of altering the frequency of shooting from distance. To answer this question, we use the reasoning techniques outlined in Section 3.2 and explore the effect of increasing and decreasing the frequency of shooting from long-distance by 5%, 10% and 20%.

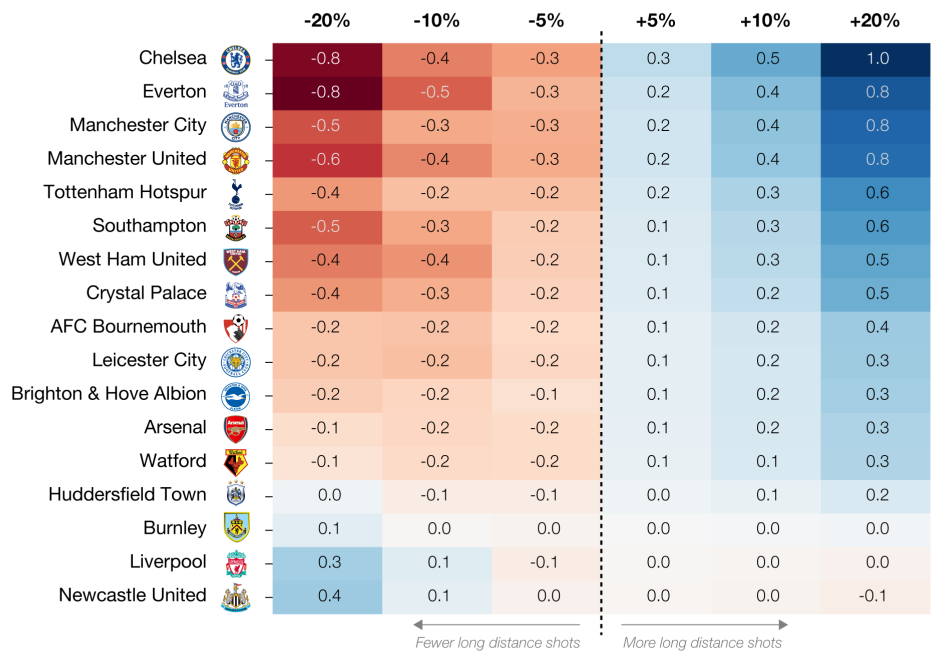

**Figure 8:** Effect on the expected number of goals a team would score when uniformly increasing or decreasing the frequency of shooting from the long-distance region shown in Figure 2 by 5%, 10% and 20%. Most teams would see an increase in the number of goals with an increase in shooting from distance. Some exceptions, like Burnley, Liverpool and Newcastle United, would see no gains or even a decrease.

Figure 8 shows the change in the expected number of goals each team would score over the course of a season as a result of altering the frequency of shooting from distance. For most teams, we see that shooting less from distance leads to a decrease in the number of goals a team would be expected to score. Increasing the number of long shots would yield more goals for most teams. The exceptions in both cases are Newcastle United, Liverpool and Burnley, who would see the opposite happening. These three teams are mainly teams with a low conversion rate for long-distance shots. Therefore, decreasing the frequency of those shots can have a positive effect. Other teams like Chelsea, Everton and Manchester City can be identified as the teams that can expect the biggest increases. All three teams have or had players with a good long-distance shot (e.g., Eden Hazard with Chelsea, Gylfi Sigurdsson with Everton, and Kevin De Bruyne with Manchester City).

---

[3] https://twitter.com/devinpleuler/status/1226919308762193920

## 4.4. Targeted increases of shots from distance

The previous use case evaluated the effect of uniformly increasing a team's propensity to shoot from all the considered long-distance locations. However, the analysis in use case 4.1 clearly illustrates that there are a limited number of team-specific long-distance zones where it may be fruitful to consider shooting more often. Therefore, in this analysis we explore the effect of each team shooting 5%, 10% or 20% more often but only from those locations where shooting was deemed to be the better choice than moving for the team.

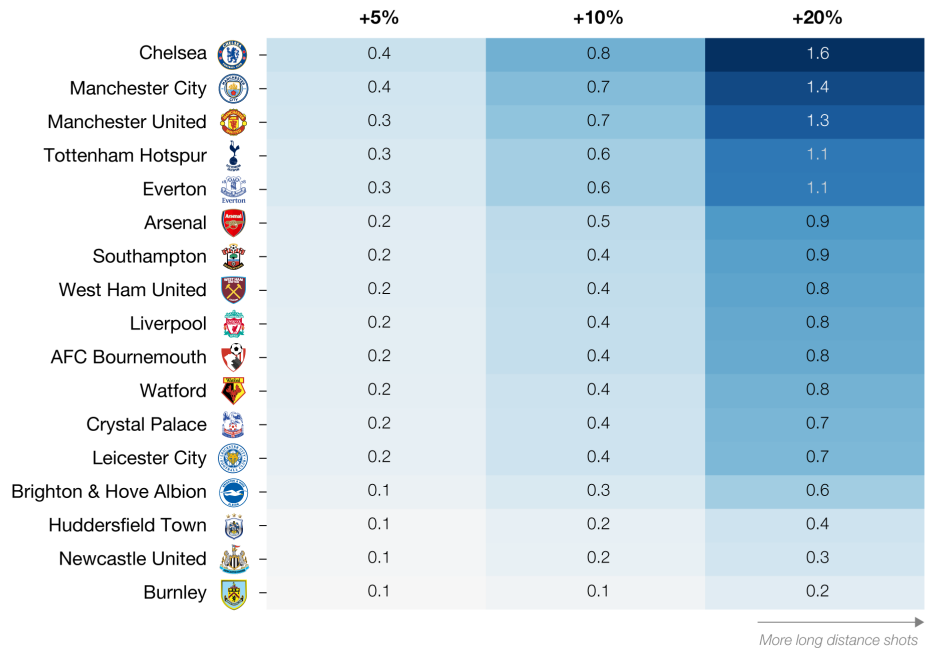

**Goal difference between a targeted adjustment and the original policy**

| Team | +5% | +10% | +20% |
|---|---|---|---|
| Chelsea | 0.4 | 0.8 | 1.6 |
| Manchester City | 0.4 | 0.7 | 1.4 |
| Manchester United | 0.3 | 0.7 | 1.3 |
| Tottenham Hotspur | 0.3 | 0.6 | 1.1 |
| Everton | 0.3 | 0.6 | 1.1 |
| Arsenal | 0.2 | 0.5 | 0.9 |
| Southampton | 0.2 | 0.4 | 0.9 |
| West Ham United | 0.2 | 0.4 | 0.8 |
| Liverpool | 0.2 | 0.4 | 0.8 |
| AFC Bournemouth | 0.2 | 0.4 | 0.8 |
| Watford | 0.2 | 0.4 | 0.8 |
| Crystal Palace | 0.2 | 0.4 | 0.7 |
| Leicester City | 0.2 | 0.4 | 0.7 |
| Brighton & Hove Albion | 0.1 | 0.3 | 0.6 |
| Huddersfield Town | 0.1 | 0.2 | 0.4 |
| Newcastle United | 0.1 | 0.2 | 0.3 |
| Burnley | 0.1 | 0.1 | 0.2 |

*More long distance shots*

**Figure 9:** Effect on the number of goals scored when performing a targeted increase in the frequency of long-distance shots. Most teams would score an extra goal, which would equate to one additional point in the league table, which could be important in terms of qualifying for the Champions League or being relegated.

Figure 9 shows the change in the expected number of goals a team would score over the course of a season. This more targeted approach now yields increases in the expected number of goals scored for all teams. Shooting 10% more often tends to yield a gain of about an additional half a goal per season. Shooting 20% more often would yield an extra goal for most teams.

Scoring an extra goal is possibly very important as for the bottom teams it could mean the difference between being relegated or staying up. For the top teams, it could mean the difference between qualifying for the Champions League or not. Relegation is a catastrophe with a cost that can be estimated at around $255 million.[4] Figure 10 shows that every goal scored during a season equates to roughly one point in the table. The figure was constructed using all data from the last ten seasons of the English Premier League (i.e., 2010/11 until 2019/20). One of the best examples of where our

---

[4] https://www.firstpost.com/sports/premier-league-250m-atstake-as-aston-villa-bournemouth-and-watford-fight-to-avoid-relegation-on-final-day-8641391.html

methods could have been useful is Bournemouth's recent relegation in the 2019/20 season, where they finished one point behind on Aston Villa. Our methods indicate that Bournemouth would have scored about one more goal, which would yield an extra point in the table and resulted in them being tied with Aston Villa. In case of ties, the goal difference (i.e., the difference between goals scored and goals conceded) is used to determine the final position. In this case, Bournemouth would then have had a goal difference of -24, whereas Aston Villa had a goal difference of -26. As a result, Aston Villa would have been relegated instead of Bournemouth.

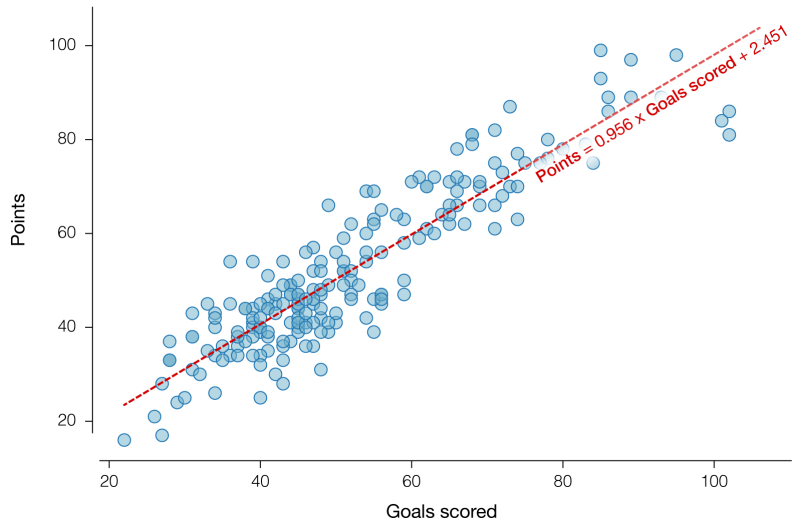

**Figure 10:** Points in the final table as a function of the goals scored. The image visualizes the data of the last ten seasons of the English Premier League (i.e., 2010/2011 until 2019/2020). Every goal scored equates to one point in the final table.

## 5. Related Work

To the best of our knowledge, the use of probabilistic model checking to compare the chances of scoring for different action sequences is novel within sports analytics. Exploring the effects of modifying a team's shooting policy has been done in basketball by Sandholtz and Bornn [14], [15]. Beyond looking at different sports, there are several important differences between our work and that of Sandholtz and Bornn. First, we adjust both the frequency and the efficiency of shooting whereas the existing work only modified the frequency of shooting. Second, we consider a more expansive action space. This action space also immediately allows for applying our methods of frequency and efficiency alterations to other decisions such as movement. In contrast, their methods modified their transition probability tensor to analyze movement decision making. Third, we compute the expected number of goals a team would score via the MDP's fundamental matrix as opposed to simulating the season. Finally, we analyze event data whereas Sandholtz and Bornn have access to tracking data which allows them to build a more fine-grained model. However, event data is generally more widespread than tracking data.

Markov Decision Processes, and Markov models in general, have been widely used in the analysis of sports such as soccer [6], [12], [13], [17], [19], American football [4], basketball [2], [14], [15], [18] and ice hockey [11], [16]. In 2011, Rudd [13] introduced the first framework for analyzing soccer using Markov Chains, and her general idea has been built upon by others [17], [19]. It is also worth noting that the work of Yam [19] uses the fundamental matrix of their Markov model to calculate the probability of a sequence resulting in a goal. Conceptually, our model is similar to these existing models. The key difference is that prior analyses using Markov models focused on assessing and rating players [12] whereas we have explored tactics and decision making.

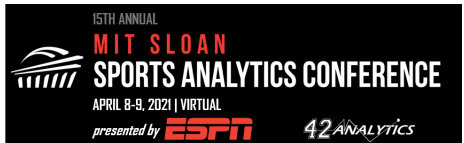

In terms of decision making and tactics, Hirotsu and Wright [6] used a Markov process to decide upon the optimal time for substitutions and formation changes. A completely different approach by Peralta et al. [9] uses self-propelled particle models to aid with player decision-making. The work of Power et al. [10] aims to bust some of the common myths around set-pieces and analyses these set-pieces in order to give recommendations to coaches for deciding upon these actions during match preparation.

## 6. Conclusions

This paper proposed a framework to reason about decision making and the possible effects of decisions in soccer by combining techniques from machine learning and artificial intelligence. Concretely, we applied this framework to explore the effects of shooting from long-distance on two different levels. First, we looked at whether shooting or moving in a location outside the penalty box would be more likely to yield a goal in the next one to two actions. We found that for each team, there are several specific areas where shooting tends to be the superior decision than attempting to move the ball in the hope of generating a better shot down the line. Second, we reasoned about what would happen if a team increased or decreased their frequency of shooting from distance over the course of a season. Our analysis indicates that teams are leaving goals on the pitch with their current shooting policies. If they were to increase their frequency of shooting from distance from a limited number of specific areas, they would be expected to score about one additional goal over the course of a season. These findings have several implications and applications for practitioners in terms of coaching players, preparing for a specific opponent, or setting a team's tactic. The proposed framework can easily be extended and applied to analyze other aspects of the game. For example, our proposed techniques can be applied to analyze whether a forward or a backward throw-in is more suitable in a specific situation, and how to effectively play out from the back. In the future, we aim to further investigate the full potential of our framework.

## Acknowledgements

We thank StatsBomb for providing the event data used in this research. We thank Jan Van Haaren and Lotte Bransen for their comments and suggestions that helped improve this paper. Maaike Van Roy, Wen-Chi Yang, Luc De Raedt and Jesse Davis are supported by the Research Foundation – Flanders under EOS No. 30992574. Pieter Robberechts and Luc De Raedt are supported by the KU Leuven Research Fund (C14/17/070 and C14/18/062, respectively). This work also received funding from the Flemish Government under the "Onderzoeksprogramma Artificiële Intelligentie (AI) Vlaanderen" programme.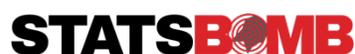

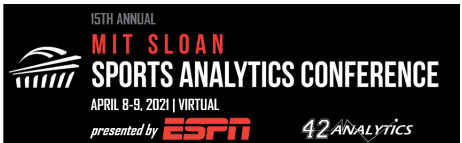

# Appendix

### A.1 Evaluation of the accuracy of our MDP

To assess how accurately our MDP models the data, we first discuss the notion of the value function of an MDP. This value function assigns a value to each state [1]:

$$V_\pi(s) = \sum_{a \in A} \pi(a \mid s) * \sum_{s' \in S} P(s, a, s') * (R(s, a, s') + \gamma * V_\pi(s'))$$

Where $S$ is used to denote the state space, $A$ is used to denote the action space, $R$ is used to denote the reward function and $\gamma$ is a discount factor. We use $\gamma = 1$, meaning that all future goals are equally important, as is the case in a real-life soccer game.

As can be seen from the formula, the value of each state depends on the values of all the other states in the MDP. Therefore, one way of solving this equation is by using dynamic programming (i.e., iteratively evaluating the value function until conversion). In soccer terms, each iteration essentially adds one more possible action to the sequence. After initializing all $V_\pi(s)$ with zero, initially the model only allows the shooting action. After one iteration, an extra possible movement action is also taken into account. After two iterations, two extra possible movement actions are taken into account, and so on [17]. Eventually, it includes all actions that are ever possible. The value of each state then coincides with the probability of eventually scoring a goal in the same possession sequence, given that the sequence initiated from that state. This probability can also be obtained by using a probabilistic model checker.

In order to assess the accuracy of our MDP, we compare the value of each game state (computed with the value function) with the empirical value of the game state obtained from the data. To compute the empirical value from the data, we count for each state the number of times an action is performed, with the condition that the action is part of a possession sequence that leads to a goal later on. To estimate the probability of eventually scoring a goal from that state, we divide this number by the total number of possession sequences passing through that state. Over all 17 teams that played in both the 2017/18 and 2018/19 English Premier League season, the mean absolute error between the results from the model and the empirical values lies in the following interval (avg $\pm$ 1std): [0.013, 0.015]. This small discrepancy can be attributed to multiple factors. Smoothing can have a small influence on the results. Additionally, teams do not demonstrate *every* possible possession sequence from each state. In contrast, the results obtained by evaluating the value function or by using a probabilistic model checker do take every possible sequence into account.

### A.2 General formulas for modifying the policy

As mentioned in the paper, when increasing (decreasing) the policy, we alter all other affected probabilities according to their original share. Mathematically, increasing the shooting probability by *x* percent can be achieved as follows:

$$\pi'(shoot \mid s) = \pi(shoot \mid s) + (1 - x) * \pi(shoot \mid s)$$

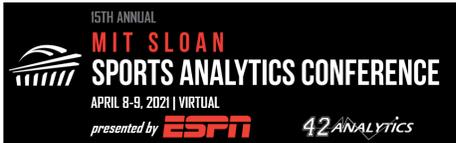

$$\forall\, s' \in field\ states{:}\ \pi'(move\_to(s') \mid s) = \frac{\pi(move\_to(s') \mid s)}{total\_move\_prob\_s} *$$

$$(total\_move\_prob\_s - (1-x) * \pi(shoot \mid s))$$

where

$$total\_move\_prob\_s = \sum_{s' \in field\ states} \pi(move\_to(s') \mid s)$$

Conversely, decreasing the shooting probability by *x* percent in a state can be realized as:

$$\pi'(shoot \mid s) = (1-x) * \pi(shoot \mid s)$$

$$\forall\, s' \in field\ states{:}\ \pi'(move\_to(s') \mid s) = \frac{\pi(move\_to(s') \mid s)}{1 - x * \pi(shoot \mid s)}$$

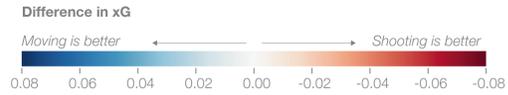

## A.3 All teams: Should you pass up on the shot, or take it?

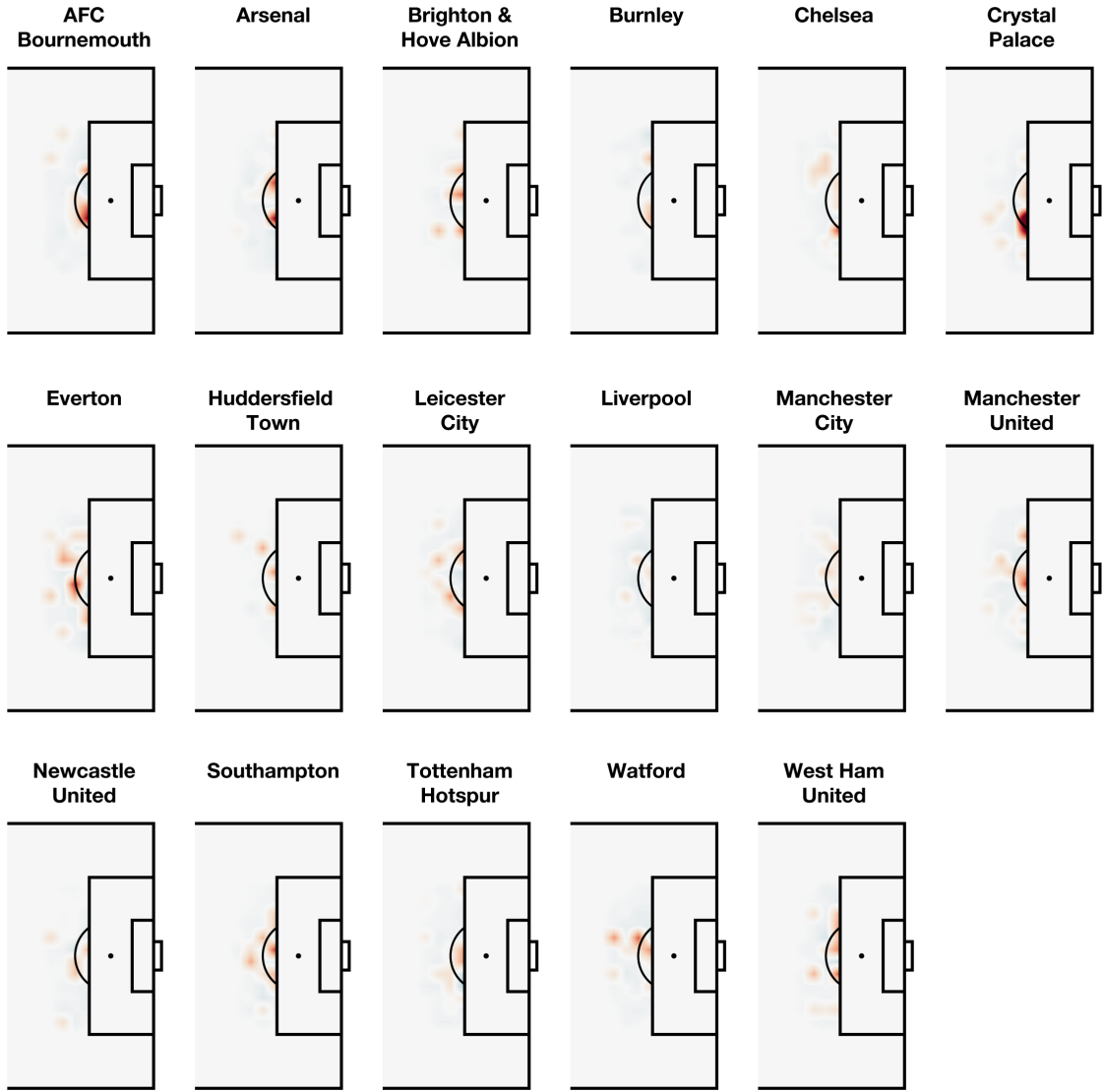

**Figure A.1:** The difference in xG between moving exactly once prior to shooting and directly shooting for all Premier League teams. Red indicates where immediately shooting is the better choice as moving would decrease the odds of scoring. Blue indicates when moving would optimize a team's chances of scoring.

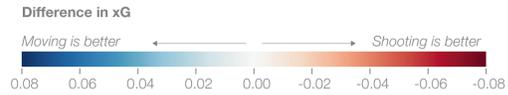
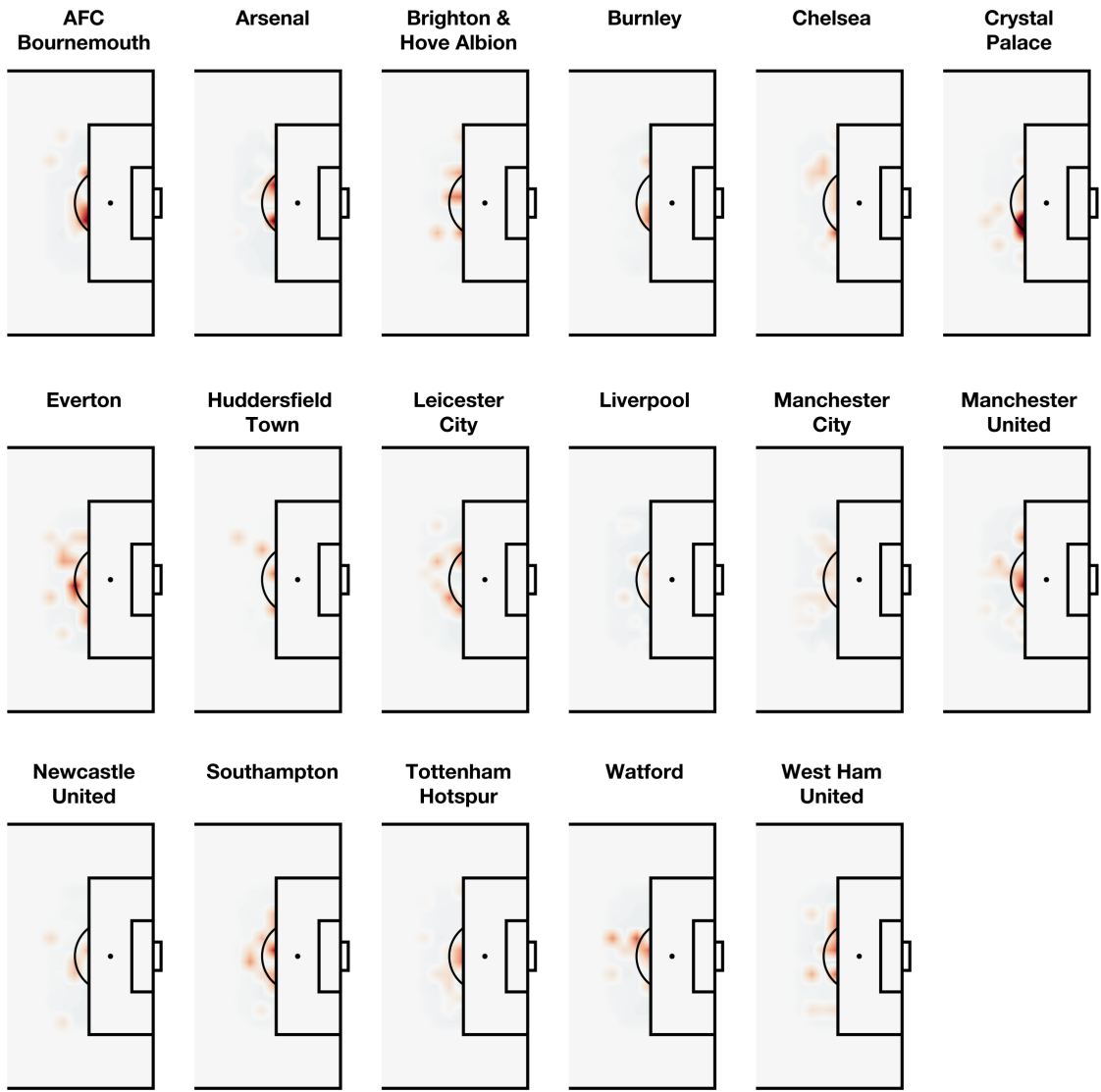

**Figure A.2:** The difference in xG between moving exactly twice prior to shooting and directly shooting for all Premier League teams. Red indicates where immediately shooting is the better choice as moving would decrease the odds of scoring. Blue indicates when moving would optimize a team's chances of scoring.

## A.4 All teams: Probability of generating a better shot

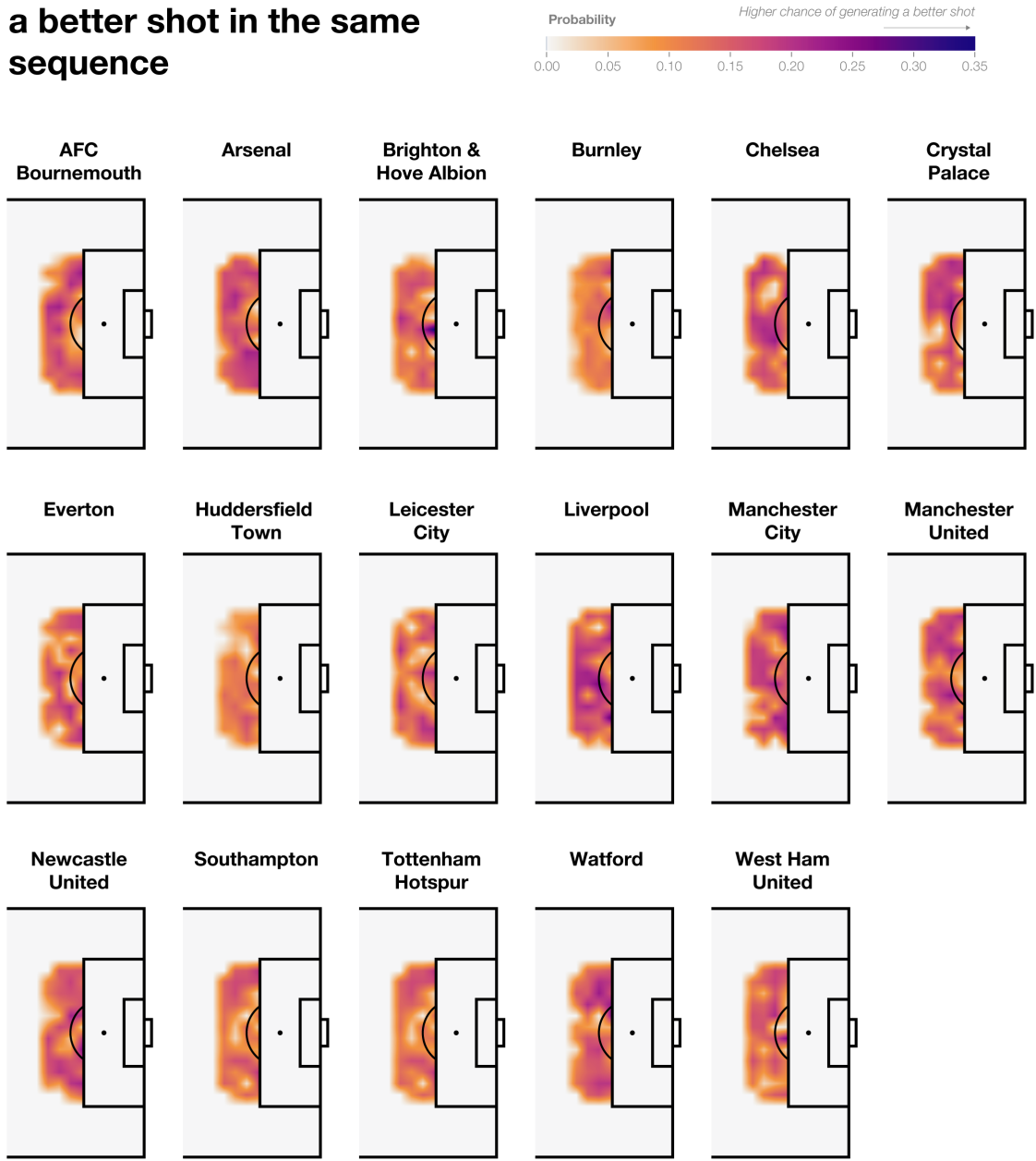

**Figure A.3:** The probability of ever generating a better shot in the same possession sequence for all Premier League teams. The regions with a low probability correspond to those regions where it is preferable for a team to shoot rather than move as determined in Figures A.1 and A.2.

25